\documentclass[10pt,twocolumn,letterpaper]{article}

\usepackage{cvpr}
\usepackage{times}
\usepackage{epsfig}
\usepackage{graphicx}
\usepackage{amsmath}
\usepackage{amssymb}

\usepackage{cvpr}
\usepackage{times}
\usepackage{epsfig}
\usepackage{graphicx}
\usepackage{amsmath}
\usepackage{amssymb}
\usepackage{booktabs}
\usepackage{algpseudocode}
\usepackage{algorithm}

\usepackage{cvpr}
\usepackage{times}
\usepackage{epsfig}
\usepackage{graphicx}
\usepackage{amsmath}
\usepackage{amssymb}

\usepackage{comment}
\usepackage[british,UKenglish,USenglish,english,american]{babel}
\usepackage{multirow}
\usepackage{algorithm}
\usepackage{algpseudocode}

\usepackage[breaklinks=true,bookmarks=false]{hyperref}

\cvprfinalcopy 

\def\cvprPaperID{****} 

\ifcvprfinal\fi
\begin{document}

\title{Pseudo Mask Augmented Object Detection}

\author{Xiangyun Zhao\thanks{This work is done when Xiangyun Zhao was an intern at Microsoft Research Asia.}\\
Northwestern University\\
{\tt\small zhaoxiangyun915@gmail.com}
\and
Shuang Liang\thanks{Corresponding author.}\\
Tongji University\\
{\tt\small shuangliang@tongji.edu.cn}
\and
Yichen Wei\\
Microsoft Research\\
{\tt\small yichenw@microsoft.com}
}
\maketitle
\thispagestyle{empty}

\begin{abstract}
In this work, we present a novel and effective framework to facilitate object detection with the instance-level segmentation information that is only supervised by bounding box annotation. Starting from the joint object detection and instance segmentation network, we propose to recursively estimate the pseudo ground-truth object masks from the instance-level object segmentation network training, and then enhance the detection network with top-down segmentation feedbacks. The pseudo ground truth mask and network parameters are optimized alternatively to mutually benefit each other. To obtain the promising pseudo masks in each iteration, we embed a graphical inference that incorporates the low-level image appearance consistency and the bounding box annotations to refine the segmentation masks predicted by the segmentation network. Our approach progressively improves the object detection performance by incorporating the detailed pixel-wise information learned from the weakly-supervised segmentation network. Extensive evaluation on the detection task in PASCAL VOC 2007 and 2012~\cite{everingham2010pascal} verifies that the proposed approach is effective.
\end{abstract}

\section{Introduction}
Recent years have seen significant progresses in object detection. Since the deep convolutional neutral network has been firstly used in R-CNN~\cite{girshick2014rich}, a lot of improvements have been made, and they improve the performance from many different aspects, \eg, deeper networks and stronger features~\cite{simonyan2015very,szegedy2015going,he2016deep}, better object proposals~\cite{ren2015faster,yang2016craft}, more discriminative and powerful features~\cite{kong2016hypernet,bell2015inside}, more accurate localization~\cite{najibi2015g,gidaris2015locnet}, focusing on a set of hard examples~\cite{lin2017focal,shrivastava2016training}.

In this work, we investigate the object detection task from another important aspect, that is, how to exploit object segmentation to improve object detection. Although it has been well recognized in the literature that the two tasks are closely related and detection could benefit from segmentation, most previous works, \eg, ~\cite{bell2015inside,dai2015instance}, share two common drawbacks.

First, they rely on accurate and pixel-wise ground truth segmentation masks for the segmentation problem. However, such mask annotation is very expensive to obtain. Instead, most large-scale object recognition datasets such as ImageNet~\cite{deng2009imagenet} and PASCAL VOC~\cite{everingham2010pascal} only provide bounding box level annotations. In addition, most of these methods only explore how to facilitate object detection with semantic image segmentation, which did not independently consider the characteristics of each instance. We argue that the instance-level segmentation task is more aligned with object detection by considering the object information from different granularity (pixel-level versus box-level). Recently, Mask-RCNN~\cite{he2017mask} unifies object detection and instance segmentation in a single network, and show that instance segmentation could help object detection. However, pixel-wise instance segmentation labeling is still required.

Second, most works have independent network structures for segmentation and detection tasks, \eg, the state-of-the-art MNC~\cite{dai2015instance} and Mask-RCNN~\cite{he2017mask}. Although the two tasks often share the same underlying convolutional features, the two networks do not directly interact with each other and the commonality between the two tasks may not be fully exploited. For the existing approaches, the benefits of jointing learning are mostly from the better learned deep feature representation as in a normal multi-task setting. It is seldom explored that how segmentation information can benefit detection directly and more closely in a deep learning framework.

In this work, we propose a novel approach that better addresses the above two issues, which augments the object detector with generated object masks from the bounding box annotation, named as Pesudo-mask Augmented Detection (PAD). It starts from a strong baseline network architecture that directly integrates the state-of-the-art Fast-RCNN~\cite{girshick2015fast} network for object detection and InstanceFCN~\cite{dai2016instance} for object segmentation, in a normal multi-task setting.

Given the baseline network, we make two major contributions. First, our PAD treats ground truth object segmentation masks as hidden variables as they are unknown, which are gradually refined by only using bounding box annotations as the supervision, called as \emph{pseudo ground truth masks}. The pseudo masks of training images and the network parameters are optimized alternatively in an EM-like way. To make the alternative learning more effective, we propose two novel techniques. Between each iteration, the pseudo masks are progressively refined by embedding a graphical model, which improves the pixel-wise estimation with a graph-cut optimization with  low-level appearance coherence and the ground truth bounding boxes as additional constraints. Beside the iteratively refined pseudo masks, we also incorporate a novel 1D box loss defined over the groundtruth box, as a supervision signal to help improve quality of pseudo masks learning, similar to LocNet~\cite{gidaris2016locnet}.

Second, based on the commonality of segmentation and detection tasks, as well as the correlations of the network structures, we propose to connect the two networks such that the segmentation information provides a top-down feedback for detection network. In this way, the learning of detection network is improved as additional supervision signals are back propagated from the segmentation branch. The top-down segmentation feedback considers two contexts, on both the \emph{the global} level and \emph{instance} level. Their effectivenesses on improving detection accuracy are both verified in experiments.

The proposed approach is validated using various state-of-the-art network architectures (VGG and ResNet) on several well-known object detection benchmarks (i.e., PASCAL VOC 2007 and 2012). The strong and state-of-the-art performance verifies its effectiveness.


\begin{figure*}[t]
	\centering
	\includegraphics[scale=0.5]{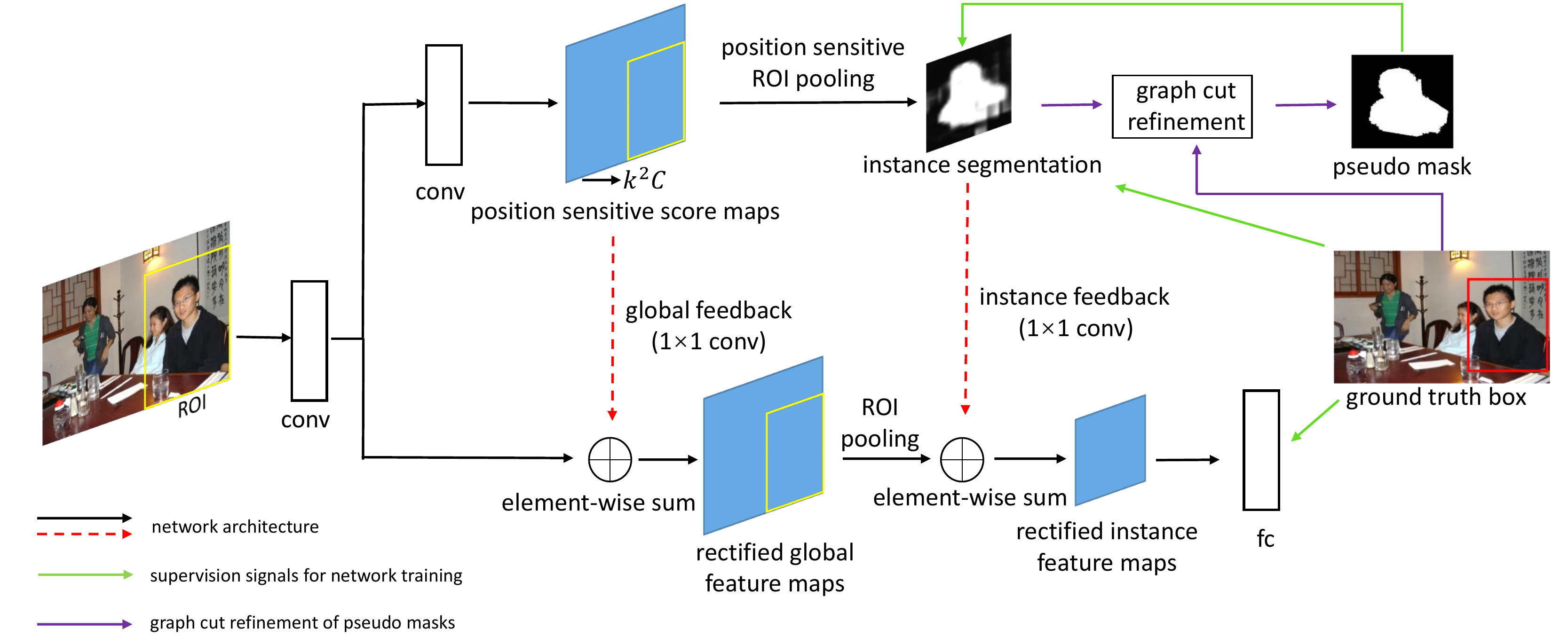}
	\caption{An overview of our pseudo-mask augmented object detection, consisting of the network architecture and graph cut based pseudo mask refinement. For each image, the detection sub-network and instance-level object segmentation sub-network share convolutional layers (i.e. conv1-conv5 for VGG, conv1-conv4 for ResNet).  For segmentation sub-network, position-sensitive score maps are generated by $1\times1$ convolutional layer, and it is then passed through position-sensitive pooling to obtain object masks. The predicted object masks and bounding box annotations are then combined to refine pseudo masks with a graph-cut refinement, which also provide the supervision signal for network training in next iteration. In each iteration, we explore the global feedback and instance top-down feedback from the instance segmentation sub-network to facilitate the object detection sub-network for better detection performance.}
	\label{fig.net1}
	\vspace{-3mm}
\end{figure*}

\section{Related Work}


\paragraph{Joint Segmentation and Detection} There exists quite a few works that integrate the object segmentation and  detection tasks~\cite{fidler2013bottom,gokberk2013segmentation,dong2014towards,chen2014enriching,bell2015inside,dai2015instance,he2017mask}. In spite of their various techniques, these methods have common limitations: 1) the pixel-level segmentation annotation is required, which is difficult to obtain, 2) the  integration of segmentation and detection is usually loose due to the separately trained segmentation and detection network. Our work overcomes the two limitations in an integrated learning framework, where the top-down segmentation feedback is proposed to bridge the segmentation and detection network.

\paragraph{Using Graphical Models for Segmentation} Graphical models are
widely used for traditional image and object segmentation ~\cite{boykov2001interactive,rother2004grabcut,li2004lazy,lafferty2001conditional,shotton2006textonboost,carreira2010constrained,koltun2011efficient}. Compared to the feature representation learning by the CNNs, the graphical inferences possess the merits of effectively incorporating local and
global image constraints (\eg appearance consistencies, and structure priors) into a single optimization framework. Recently, some recent works integrate graphical models (e.g., CRF/MRF) into the deep neutral networks~\cite{zheng2015conditional,liu2015semantic} for a joint training.


In our approach, traditional graph cut based optimization~\cite{boykov2001interactive} is embedded to refine the pseudo ground truth mask estimation during the iterative learning. It effectively refines the quality of pixel-wise pseudo masks to progressively improve the discriminative capability of detection and segmentation network.

\paragraph{Weakly Supervised Segmentation} Due to the difficulty of obtaining large-scale pixel-wise segmentation ground truth, some works resort to weakly supervised learning of segmentation, such as using bounding box annotation~\cite{dai2015boxsup, khoreva2017simple} or scribbles~\cite{lin2016scribblesup}. Such methods share some similarity with ours by using the iterative optimization to gradually refine the segmentation. They mainly focus on single image segmentation, while our approach jointly optimizes the detection and weakly-supervised object segmentation network. \cite{dai2015boxsup} does not adopt low-level color information to refine the segmentation. The most relevant work is~\cite{khoreva2017simple}. It also iteratively refine the segmentation by graphical models (CRF). Different from it, our approach aims to improve object detection with weakly supervised segmentation.

\section{Pseudo-mask Augmented Detection}
We focus on facilitating the object detection with the instance-level object segmentation information, using only ground truth bounding box annotations. We denote the set of all ground truth boxes as $B^{gt}=\{B^{gt}_o\}$, where subscript $o$ enumerates all objects in all training images. We use the former notation throughout the paper for its simplicity.

As motivated earlier, it is beneficial to estimate the per-pixel object segmentation as well. An auxiliary object segmentation task is added in a normal multi-task setting. That is, the two tasks share the same underlying convolutional feature maps. Since the ground truth binary object segmentation masks are unknown, we treat them as hidden variables, which are first initialized with $B^{gt}$, and then iteratively refined in our approach. We call them estimated object masks as \emph{pseudo ground truth masks} from the bounding box annotation, denoted as $M^{pseudo}=\{M^{pseudo}_o\}$.

Let the network parameters be $\Theta$, and the network output for object segmentation and detection be $M(\Theta)$ and $B(\Theta)$, respectively. The network parameters are learned to minimize the loss function

\begin{equation}
L_{seg}(M(\Theta)|M^{pseudo},B^{gt}) + L_{det}(B(\Theta)|B^{gt}),
\label{eq.loss_network_loss}
\end{equation}
where the two loss terms are enforced on object segmentation and detection tasks, respectively. As defined the network optimization target, the performance of detection network heavily depends on the quality of estimated pseudo masks $M^{pseudo}$. That is, the poor estimation of $M^{pseudo}$ leads to poor network learning of $M(\Theta)$, which in turn would cause negative chain effect on whole iterative framework for object detection. We propose an effective learning approach that progressively improves the quality $M^{pseudo}$ from a coarse initialization using $B^{gt}$, as summarized in Algorithm~\ref{alg.training}. The detection network parameters $\Theta$ and pseudo masks $M^{pseudo}$ are alternatively optimized following a EM-like way, with the other fixed in each iteration.

Note that Algorithm~\ref{alg.training} only operates on training images. The learned network parameters $\Theta$ are applied on test images to generate detection and segmentation results.

The instance-level segmentation masks $M(\Theta)$ from the pixel-wise prediction of segmentation network are usually noisy and poor. This is partially because pseudo masks are not accurate enough, and the estimation is made in a pixel-wise manner, which does not consider the correlations between the pixels such as smoothness constraints used in most segmentation approaches. As shown in Algorithm~\ref{alg.training}, we thus propose two novel ingredients to achieve the effective iterative learning. First, in each object mask refinement step (Sec.~\ref{sec.pseudo_mask_refinement}), the pseudo ground truth mask for each object is improved using the traditional graphical inference. It is formulated as a global optimization problem that considers not only the current mask estimation from the network, but also the low level image appearance coherence and the ground truth bounding boxes, which is efficiently solved by graph cut~\cite{boykov2004graphcut}.

Second, we notice that only using the pseudo mask $M^{pseudo}$ as 2D pixel-wise supervision signals may be not sufficient as the masks themselves are often noisy and not accurate enough. Thus, the 1D box loss( explained in Sec.~\ref{sec.multi_task_net}) in Eq.~\eqref{eq.loss_network_loss} and ~\eqref{eq:loss_seg} (Sec.~\ref{sec.multi_task_net}) is incorporated to consider the additional constraints provided by the ground truth bounding box. The 1D loss term complements the noisy 2D segmentation loss and performs better regularization on the segmentation network learning.

With the aforementioned two novel techniques, both pseudo masks $M^{pseudo}$ and network parameters $\Theta$ are improved steadily, benefiting from each other. Based on the refined object masks, we add connections between the segmentation and detection sub-networks such that the segmentation features provide top-down feed back for the detection, leading to better results in the object detection (Sec.~\ref{sec.multi_task_net}).

\begin{algorithm}[t]
\caption{Iterative learning of network parameters $\Theta$ and pseudo ground truth masks $M^{pseudo}$.}
\begin{algorithmic}[1] 
\State \textbf{input}: ground truth bounding boxes $B^{gt}$
\State initialize the pseudo masks $M^{pseudo}$ from $B^{gt}$;
\State learn $\Theta_0$ with loss in Eq.~(\ref{eq.loss_network_loss})                 \Comment{Sec.~\ref{sec.multi_task_net}}

\For{$t=1$ \textbf{to} $T$}                                  
\State refine $M^{pseudo}$ from $M(\Theta_{t-1})$ and $B^{gt}$                      \Comment{Sec.~\ref{sec.pseudo_mask_refinement}}
\State learn $\Theta_t$ with loss in Eq.~(\ref{eq.loss_network_loss})                 \Comment{Sec.~\ref{sec.multi_task_net}}
\EndFor

\State \textbf{output}: final network parameters $\Theta_t$
\State \textbf{output}: pseudo ground truth masks $M^{pseudo}$
\end{algorithmic}
\label{alg.training}

\end{algorithm}
\begin{figure*}[t]
	\centering
	\includegraphics[scale=0.5]{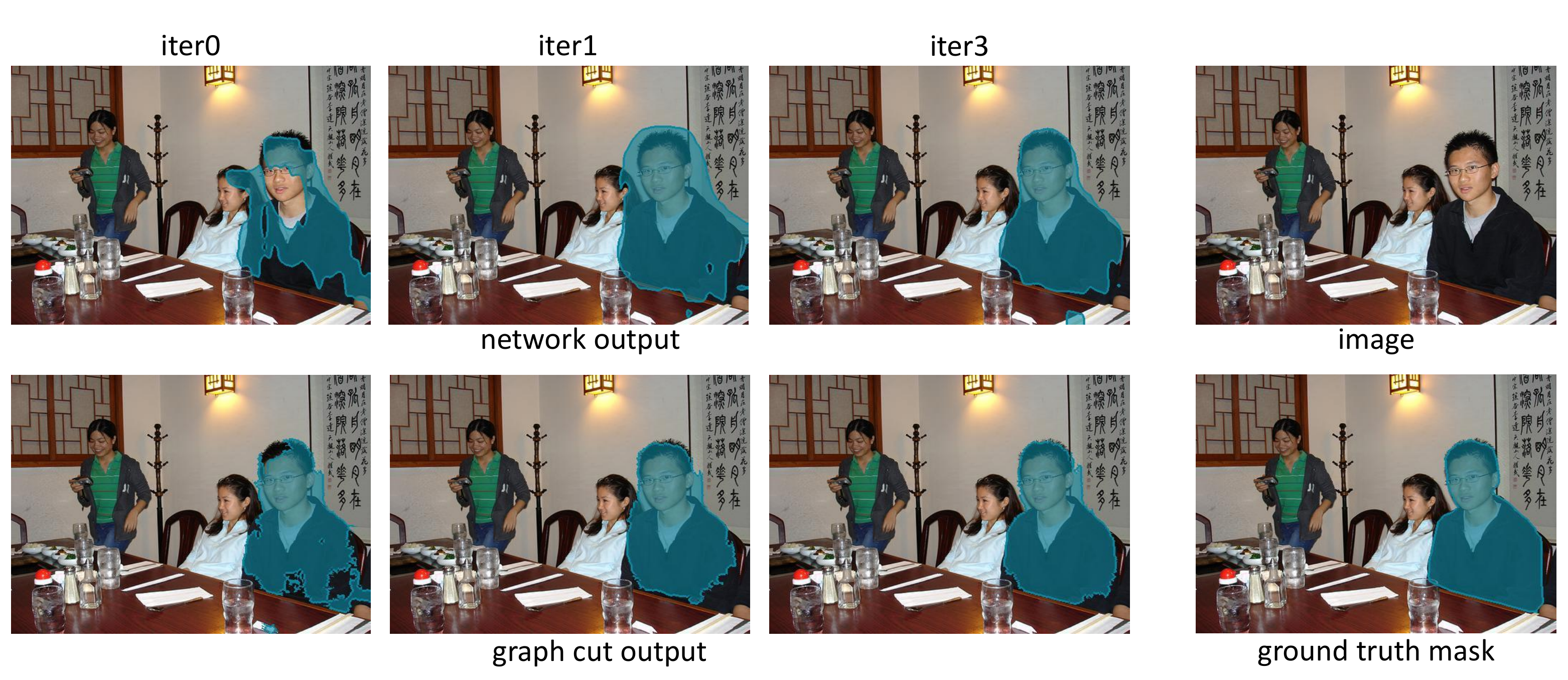}\vspace{-2mm}
	\caption{Example predicted pseudo masks by the iterative refinement and employing graph cut optimization.}
	\label{fig.graph_cut_example_result}\vspace{-4mm}
\end{figure*}

\subsection{Network Architecture and Training}
\label{sec.multi_task_net}

As shown in Figure~\ref{fig.net1}, following the common multi-task learning, we adopt two sub-networks for the object segmentation and detection tasks, which are built on the shared feature maps. We first extract the object proposals, or regions of interest (ROIs) from the Region Proposal Network (RPN)~\cite{ren2015faster}. For simplicity, we do not use the complex training strategy in~\cite{ren2015faster}. Instead, we pre-train the RPN and fix the ROIs throughout our experiments.

\paragraph{Object Segmentation with Pseudo Masks}
In general, this sub-network can adopt any instance-level object segmentation network such as DeepMask~\cite{pinheiro2015learning}, MNC~\cite{dai2015instance}, etc. In this work, we adopt the similar technique as in InstanceFCN~\cite{dai2016instance} and FCIS~\cite{li2017fully}, which are state-of-the-art methods for instance segmentation. It applies $1 \times 1$ convolutional layer on the feature maps to produce $k^2$ \emph{position sensitive} score maps. The $k^2$ (we use $k=7$ in this work) maps encode the relative-position information between image pixels and ROIs (\eg, top-left, bottom-right). Given a ROI, its per-pixel score map is assembled by dividing the positive sensitive score maps into $k^2$ cells and copying each cell's content from the corresponding $k^2$ maps. This step generates a fixed-size ($28\times28$) per-ROI foreground probability map.

However, this strategy still has several limitations: first, it only runs on square sliding windows; second, it ignores object category information and is limited to object generate object segment proposals. We extend this approach in two aspects to seamlessly integrate it into the object detection network. First, we extend its $k^2$ score maps to $k^2*C$ score maps, where $C$ indicates the category number. In this way, the individual segmentation module for each category is optimized. Second, we employ generic object proposals~\cite{ren2015faster} to replace the square sliding windows and the position-sensitive ROI pooling layer in~\cite{dai2016rfcn} on the proposals. During training, a sigmoid layer is applied on each category of the per-ROI score maps to generate the instance foreground probability maps. 


\paragraph{Segmentation Loss}
Our approach estimate a corresponding pseudo mask for each object instance. For a ROI that has intersection-over-union (IOU) larger than $0.5$ with a ground-truth object, we define a per-pixel sigmoid cross-entropy loss on the ROI's foreground probability map, with respect to current pseudo mask of the object that is regarded as a hidden ground truth mask. We call this term \emph{2D mask loss}.

Since the pseudo masks are quite noisy, it may damage our network if directly using it as the supervision. Since each ground truth bounding box tightly encloses the object, this implies that for a horizontal or vertical scan line in the box, at least one pixel on the line should be foreground. On the other hand, all pixels outside of the box should be background. Accordingly, we define a \emph{1D box loss} term for each ROI. Specifically, the predicted foreground mask of each ROI is projected to two 1-d vectors along the horizontal and vertical directions, respectively, by applying a max operation on all values of a scan line. For each position on the 1-d vector, it is denoted as foreground when its corresponding line is inside the box. Otherwise, it is denoted as background. The 1D loss term summarizes the sigmoid cross-entropy loss on all positions of the 1-d vectors. Note that a similar 1D loss idea has also been utilized in LocNet~\cite{gidaris2016locnet} for bounding box regression. In this work, we use it for object segmentation.

In summary, by combining the 2D mask loss and 1D box loss, the segmentation loss in Eq.~\eqref{eq.loss_network_loss} is computed by

\begin{equation}
L_{\rm seg} = L_{2D}(M(\Theta)|M^{pseudo}) + L_{1D}(M(\Theta)|B^{gt}).
\label{eq:loss_seg}
\end{equation}

\paragraph{Object Detection with Top-down Segmentation Feedback}
For the detection sub-network, we use the state-of-the-art Fast(er) R-CNN~\cite{girshick2015fast,ren2015faster}. It applies a ROI pooling layer for each ROI on the feature maps to obtain per-ROI feature maps, and then applies fully connected (FC) layers to output detection results.

In the common multi-task setting, the two sub-networks do not interact and only use the separate optimization target. However, the object segmentation and detection tasks are highly correlated to each other and the sub-networks also share similar structures, we connect the two so that the segmentation network provides top-down feedback information for detection network.

The feedback from segmentation consists of the \emph{global feedback} and the \emph{instance feedback} (the two red dotted arrows in Figure~\ref{fig.net1}).  In terms of the global level, the $k^2*C$ position-sensitive score maps in the segmentation sub-network (before ROI pooling) encode the segmentation information on the whole image. They are of the same spatial dimension of the shared convolutional feature maps but different feature channels, in general. A $1\times1$ convolutional layer is applied on the score maps to change its channel number to match that of the shared feature maps. The two sets of maps are then element-wisely summed to produce the ``rectified" global feature maps for the object detection sub-network.

In terms of instance level, the instance segmentation masks (i.e., per-ROI score maps) encode the specific pixel-wise characteristics of each object instances. As shown in Figure~\ref{fig.net1}, after the ROI pooling step in both sub-networks, the per-ROI instance segmentation score maps from the segmentation branch are passed through a $1\times 1$ convolutional layer and max pooling layer to obtain feature maps with the same dimension as the per-ROI feature maps of the detection branch. The score maps from two branches are then summed to produce the ``rectified'' instance feature maps.

Afterwards, several fully connected (FC) layers are used to generate the object classification scores and bounding box regression results, in the same way as Fast RCNN~\cite{girshick2015fast}. The detection loss in Eq.~\eqref{eq.loss_network_loss} includes the classification soft-max loss and bounding box regression loss for all ROIs,
\begin{equation}
L_{det} = L_{cls}(B(\Theta)|B^{gt}) + L_{reg}(B(\Theta)|B^{gt}).
\label{eq.loss_det}
\end{equation}

\paragraph{Training}

Given the estimated pseudo masks $M_{\rm pseudo}$ in each step, the instance-level segmentation network and object detection network are optimized by stochastic gradient descent, using image centric sampling~\cite{girshick2015fast}. In each mini-batch, two training images are randomly sampled. The loss gradients from Eq.~\eqref{eq.loss_network_loss} are back propagated to update all the network parameters jointly.


\subsection{Pseudo Mask Refinement}
\label{sec.pseudo_mask_refinement}
Accurate pseudo mask is the key to bridge the object detection and instance-level segmentation networks. The estimated pseudo masks directly from the segmentation network are usually noisy and blurred. More importantly, as the pixels are considered individually by the convolutional network, the informative interactions between pixels are not fully exploited, such as the smoothness constraint used in traditional image and object segmentation.
\begin{table*}
	\begin{center}
		\begin{tabular}{l|c|c|c|c|c|c|c}
			\hline
			method & \footnotesize w/ seg. loss & \footnotesize w/ mask refinement & \footnotesize w/ global feedback & \footnotesize w/ instance feedback & \footnotesize VGG-16 & \footnotesize ResNet-50 & \footnotesize ResNet-101 \\
			\hline\hline
			
			ION~\cite{bell2015inside} &            &            &            &            & 75.6 &  &  \\
			CRAFT~\cite{yang2016craft} &            &            &            &            & 75.7 &  &  \\
			HyperNet~\cite{kong2016hypernet} &            &            &            &            & 76.3 &  & \\
            RON~\cite{kong2017ron} &            &            &            &            & 75.4 &  & \\
			Faster-RCNN~\cite{ren2015faster} &            &            &            &            & 73.2 & 74.9 & 76.4 \\
\hline
			Ours (a) &            &            &            &            & 74.5 & 78.1 & 79.4 \\
			Ours (b) & \checkmark &            &            &            & 74.9 & 78.2 & 79.6 \\
			Ours (c) & \checkmark & \checkmark &            &            & 75.7 & 78.6 & 79.9 \\
			Ours (d) & \checkmark & \checkmark & \checkmark &            & 76.0 & 78.7 & 80.0 \\
			Ours (e) & \checkmark & \checkmark &            & \checkmark & 76.2 & 78.9 & 80.4 \\
			Ours (f) \scriptsize iter. 1 & \checkmark & \checkmark & \checkmark & \checkmark & 75.9 & 78.9 & 80.1 \\
			Ours (g) \scriptsize iter. 2 & \checkmark & \checkmark & \checkmark & \checkmark & 76.4 & 79.2 & 80.4 \\
			Ours (h)  \scriptsize iter. 3 & \checkmark & \checkmark & \checkmark & \checkmark & \textbf{77.0} & \textbf{79.6} & \textbf{80.7} \\
			\hline
		\end{tabular}
	\end{center}
	\caption{Object detection results on VOC 2007 test training on the union set of VOC 2007, VOC 2012 train and validation dataset}
	\label{tab:ablation}
\vspace{-4mm}
\end{table*}

In this work, we explore a graphical model to refine the pseudo mask estimation, which jointly incorporates the current mask probabilities from the instance-level segmentation network, the low level image appearance cues and the ground-truth bounding box information. The graphical model is defined on a graph constructed by the super-pixels generated by~\cite{felzenszwalb2004efficient} for each object instance. For each graph, a vertex denotes a super-pixel while an edge is defined over neighboring super-pixels. Note that in this step the spatial range of the pseudo mask is enlarged by 20\% from the ground truth bounding box, in order to include more boundary areas and thus improve segmentation quality.

Formally, for all super-pixels $\{x_i\}$ in the pseudo mask under consideration, we estimate their binary labels $\{y_i\}$, where $y_i=1$ indicates foreground, and 0 for background. Similar to the traditional object segmentation approaches~\cite{boykov2001interactive,rother2004grabcut}, we define a global objective function in the form of

\begin{equation}
\sum_i U(y_i) + \sum_{i,j} V(y_i,y_j).
\label{eq:loss_graph_cut}
\end{equation}

\textbf{Unary Term.} The unary term $U(y_i)$ measures the likelihood of the super pixel $x_i$ being foreground. It considers both the foreground probabilities from the network and the ground truth bounding box $b^{gt}$, defined as

\begin{equation}
U(y_i) =
\begin{cases}
0				    & \text{if } y_i = 0 \text{ and } x_i \notin b^{gt} \\
\infty  		    & \text{if } y_i = 1 \text{ and } x_i \notin b^{gt} \\
-log(1-prob_{fg}(x_i))	& \text{if } y_i = 0 \text{ and } x_i \in b^{gt} \\
-log(prob_{fg}(x_i))  	& \text{if } y_i = 1 \text{ and } x_i \in b^{gt}. \\
\end{cases}
\end{equation}

The first two cases ensure that $x_i$ is background when it is outside the ground truth bounding box. The last two cases directly adopt the results from the current network estimation when $x_i$ is inside, where $log(prob_{fg}(x_i))$ is simply the summation of the pixel-wise
log probability of all pixels in the super-pixel $x_i$. To obtain a pixel's foreground probability, firstly we only consider the probability of the ground truth object category as the foreground probability. Secondly, the segmentation sub-network outputs all ROIs' mask probability maps. To obtain the foreground probability on a single ground truth object, we find all ROIs with IoU larger than 0.5 with the ground-truth object, and average their foreground probabilities together as the foreground probabilities for the object.

\textbf{Pairwise Term.} The pair-wise binary term $V(y_i,y_j)$ considers the local smoothness constraints, defined on all neighboring super pixels. It uses the low level image cues similarly as in~\cite{boykov2001interactive,rother2004grabcut}. If the neighboring super-pixels are similar in appearance, the cost of assigning them different labels should be high. Otherwise, the cost is low.

We use both color and texture information to measure the similarity as in \cite{lin2016scribblesup}. For a super-pixel $x_i$, its color histogram $h_c(x_i)$ is built on the RGB color space using 25 bins for each channel. The texture histogram $h_t(x_i)$ is built on the gradients at the horizontal and vertical orientations with 10 bins for each. The pair-wise binary term is defined as

\begin{equation}
\begin{aligned}
V(y_i, y_j) = & [y_i \neq y_j] \Big\{-\frac{\|h_c(x_i) - h_c(x_j)\|^2_2}{\delta^2_c}\\
& - \frac{\|h_t(x_i) - h_t(x_j)\|^2_2}{\delta^2_t} \Big\},
\end{aligned}
\end{equation}
where $[\cdot]$ is 1 or 0 if the subsequent argument is true/false. $\delta_c$ and $\delta_t$ are set as 5 and 10, respectively.

The objective function in Eq.~\eqref{eq:loss_graph_cut} is minimized by graph cut solver~\cite{boykov2004graphcut}, for the pseudo mask of each object instance in each image. The resulting binary labels $\{y_i\}$ define the refined binary pseudo masks $M_{pseudo}$, which are then used as supervision signals to train the network in next iteration (Algorithm~\ref{alg.training}).Some exemplar predicted pseudo masks of object instances are illustrated in Figure~\ref{fig.graph_cut_example_result}. 

\section{Experiments}
\subsection{Implementation and Training Details}
Our experiments are based on Caffe~\cite{jia2014caffe} and public Faster-RCNN code~\cite{ren2015faster}. For simplicity, the region proposal network (RPN) is trained once and the obtained object proposals are fixed. We evaluate the performance of the proposed PAD using three state-of-the-art network structures: VGG-16~\cite{simonyan2014very}, ResNet-50~\cite{he2016deep} and ResNet-101~\cite{he2016deep}. We use the publicly available pre-trained model on ILSVRC2012~\cite{russakovsky2015imagenet} to initialize all network parameters. 

Each mini-batch contains 2 randomly selected images, and we sample 64 region proposals per image leading to 128 ROIs for each network updating step.
After training the baseline Faster R-CNN model~\cite{ren2015faster} with OHEM~\cite{shrivastava2016training} using the above settings, we actually obtain better accuracy than that reported in the original Faster R-CNN~\cite{ren2015faster}, as also revealed in Table~\ref{tab:ablation} . 

We run SGD for 80k iterations with learning rate 0.001 and 40k iterations with learning rate 0.0001. The iteration number $T$ in Algorithm~\ref{alg.training} is set as 3 since no further performance increase is observed.

\subsection{Ablation study on VOC 2007}

Table~\ref{tab:ablation} compares different strategies and variants in our proposed approach, as well as the results from representative state-of-the-art works as reference. Following the protocol in~\cite{girshick2015fast}, all models are trained on the union set of VOC 2007~\cite{everingham2010pascal} \emph{trainval} and VOC 2012 \emph{trainval}, and are evaluated on VOC 2007 \emph{test} set. We evaluate results using VGG-16~\cite{simonyan2015very}, ResNet-50~\cite{he2016deep}, and ResNet-101 models.

We start from the baseline where no pseudo mask is used (Table~\ref{tab:ablation}(a)). This is equivalent to our faster R-CNN implementation, which sets a strong and clean baseline. It achieves 74.5\%, 78.1\%, and 79.4\% mAP scores by using VGG-16, ResNet-50, and ResNet-101 models, respectively. We then evaluate the naive pseudo mask baseline (Table~\ref{tab:ablation}(b)). 
This is equivalent to a simple multi-task baseline with coarse pseudo masks. It obtains slightly higher accuracies than (a). It indicates that multi-task learning is slightly helpful but limited, as pseudo mask quality is very poor.

To evaluate iterative mask refinement, we report the results of our variant that does not use the mask feedback for object detection network, as Table~\ref{tab:ablation}(c). It differs from  Table~\ref{tab:ablation}(b) in that the iterative mask refinement is used. After three iterations, the mAP scores are 75.7\%, 78.6\%, and 79.9\%, respectively, which are 1.2\%, 0.5\%, and 0.5\% higher than Table~\ref{tab:ablation}(a). It verifies that our approach is capable of generating more reasonable pseudo masks.

Furthermore, after performing the top-down segmentation feedback, the detection performance can be further improved by comparing the results of Table~\ref{tab:ablation}(h) and Table~\ref{tab:ablation}(c). As shown in Table~\ref{tab:ablation}(f), Table~\ref{tab:ablation}(g) and Table~\ref{tab:ablation}(h), the detection performance improve steadily over iterations. Our full model achieves mAP scores of 77.0\%, 79.6\%, and 80.7\% using different networks, respectively, which are 2.5\%, 1.5\%, and 1.3\% higher than Table~\ref{tab:ablation}(a). 
To disentangle the effectiveness of the global and instance feedbacks, we block one of them respectively in Table~\ref{tab:ablation}(d) and Table~\ref{tab:ablation}(e). We observe that both the feedbacks are effective for boosting the object detection accuracy, and combining them achieves the largest gain.

\subsection{Detection Results on VOC 2012}
The training data is the union of VOC 2007, VOC 2012 train and validation dataset, following~\cite{girshick2015fast}. As reported in Table~\ref{tab:voc_2012_test},\footnote{\url{http://host.robots.ox.ac.uk:8080/anonymous/PTJSWL.html}, \url{http://host.robots.ox.ac.uk:8080/anonymous/KZDLBX.html}}, our approach obtains 74.4\% and 79.5\% with VGG-16 and ResNet-101, which are substantially better than currently leading methods.

\setlength{\tabcolsep}{5pt}
\renewcommand{\arraystretch}{1.3}
\begin{table}[!t]
	\begin{tabular}{%
			@{\hskip 0.2em}p{4.2cm}
			@{\hskip 0.2em}c@{\hskip 0.5em}|
			@{\hskip 0.2em}c@{\hskip 0.5em}|
			@{\hskip 0.2em}c
		}
		\toprule
		\textbf{Method} &
		\textbf{Train} &
		
		\textbf{mAP} &  \\
		\midrule
		HyperNet~\cite{kong2016hypernet}  &07++12 & 71.4
		\\
         RON~\cite{kong2017ron}  &07+12 & 73.0
		\\       
		CRAFT~\cite{yang2016craft}  &07++12 & 71.3
		\\
		MR-CNN~\cite{gidaris2015object}  &07++12 &73.9
		\\
		Faster-RCNN(VGG16)~\cite{ren2015faster}  &07++12 &70.4
		\\
		Faster-RCNN(ResNet100)&07++12  &73.8
		\\
\hline
		PAD (VGG-16) &07+12 & \textbf{74.4}
		\\
		PAD (ResNet-101)&07+12 & \textbf{79.5}
		\\
		\bottomrule
	\end{tabular}
	\caption{Detection results on VOC 2012 test. 07+12: 07 trainval + 12 trainval, 07++12: 07 trainvaltest + 12 trainval.}
	\label{tab:voc_2012_test}\vspace{-4mm}
\end{table}

\setlength{\tabcolsep}{2pt}
\renewcommand{\arraystretch}{1.3}
\begin{table}
	\begin{center}
		\begin{tabular}{l|c|c|c}
			\hline
			method & train w/ gt mask? & mAP$^r$ (\%) & mAP$^b$ (\%) \\
			\hline\hline
			\hline
			Faster R-CNN & & - & 66.3 \\
			\hline
			MNC~\cite{dai2015instance} & \checkmark & 63.5 & - \\
			\hline
			PAD \scriptsize w/ gt mask & \checkmark & \textbf{64.5} & \textbf{68.1} \\
			\hline
			PAD \scriptsize w/ Grabcut mask & & 48.3 & 66.9 \\
			PAD \scriptsize w/o 1D box loss & & 49.1 & 66.9 \\
			\hline
			PAD \scriptsize iter. 0 & & 44.3 & 66.7 \\
			PAD \scriptsize iter. 1 & & 52.1 & 67.0 \\
			PAD \scriptsize iter. 2 & & 58.0 & 67.5 \\
			\textbf{PAD} & & \underline{58.5} & \underline{67.6} \\
			\hline
		\end{tabular}
	\end{center}
	\caption{Performance comparison on VOC 2012 SDS task.}
	\label{tab:voc_sds}\vspace{-4mm}
\end{table}

\begin{figure*}[t]
	\centering
	\includegraphics[scale=0.47]{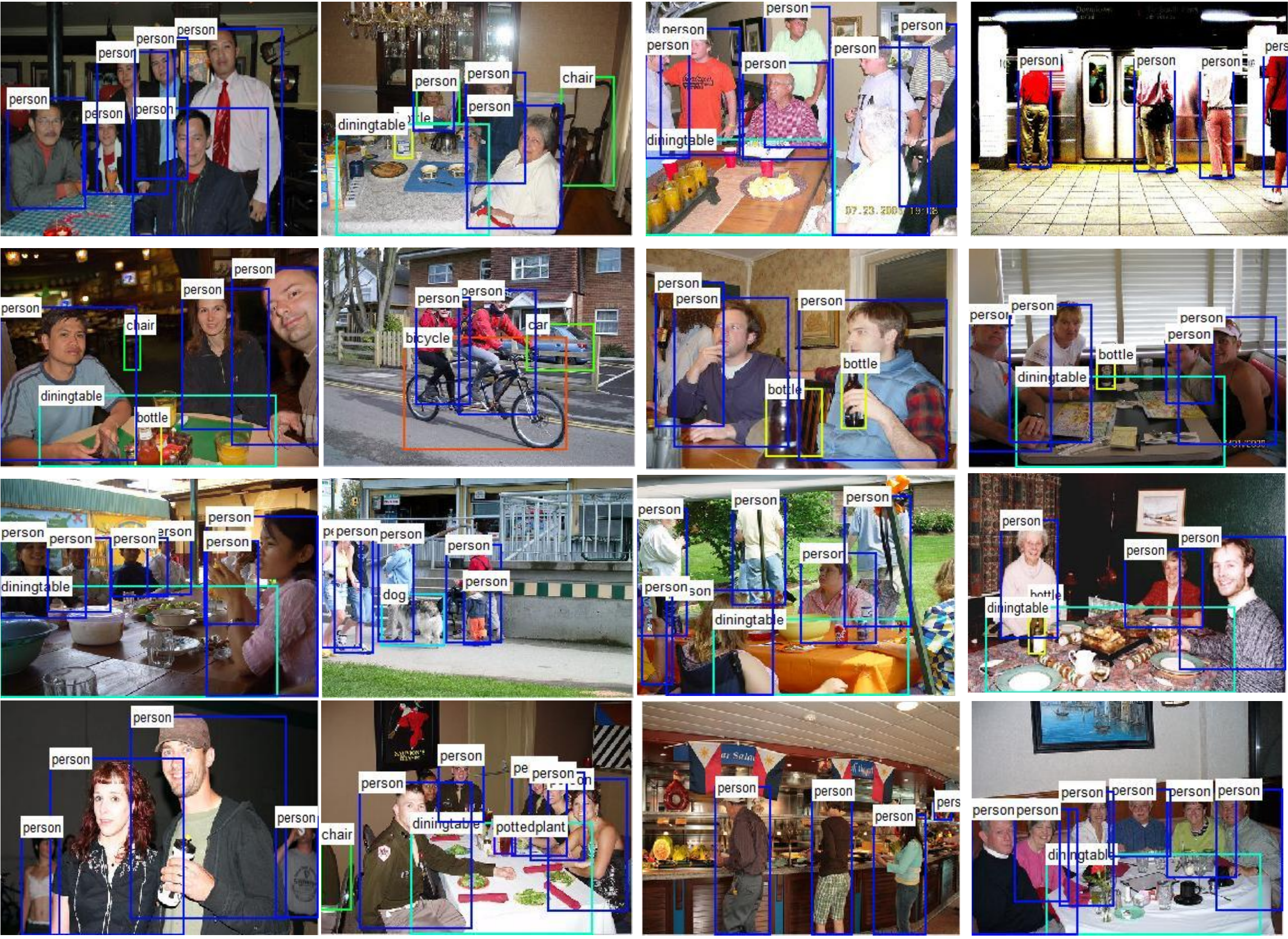}\vspace{-2mm}
              \caption{Example object detection results of our approach SDS validation set}\vspace{-4mm}
              \label{fig.test_det}
\end{figure*}

\begin{figure*}[t]
	\centering
	\includegraphics[scale=0.52]{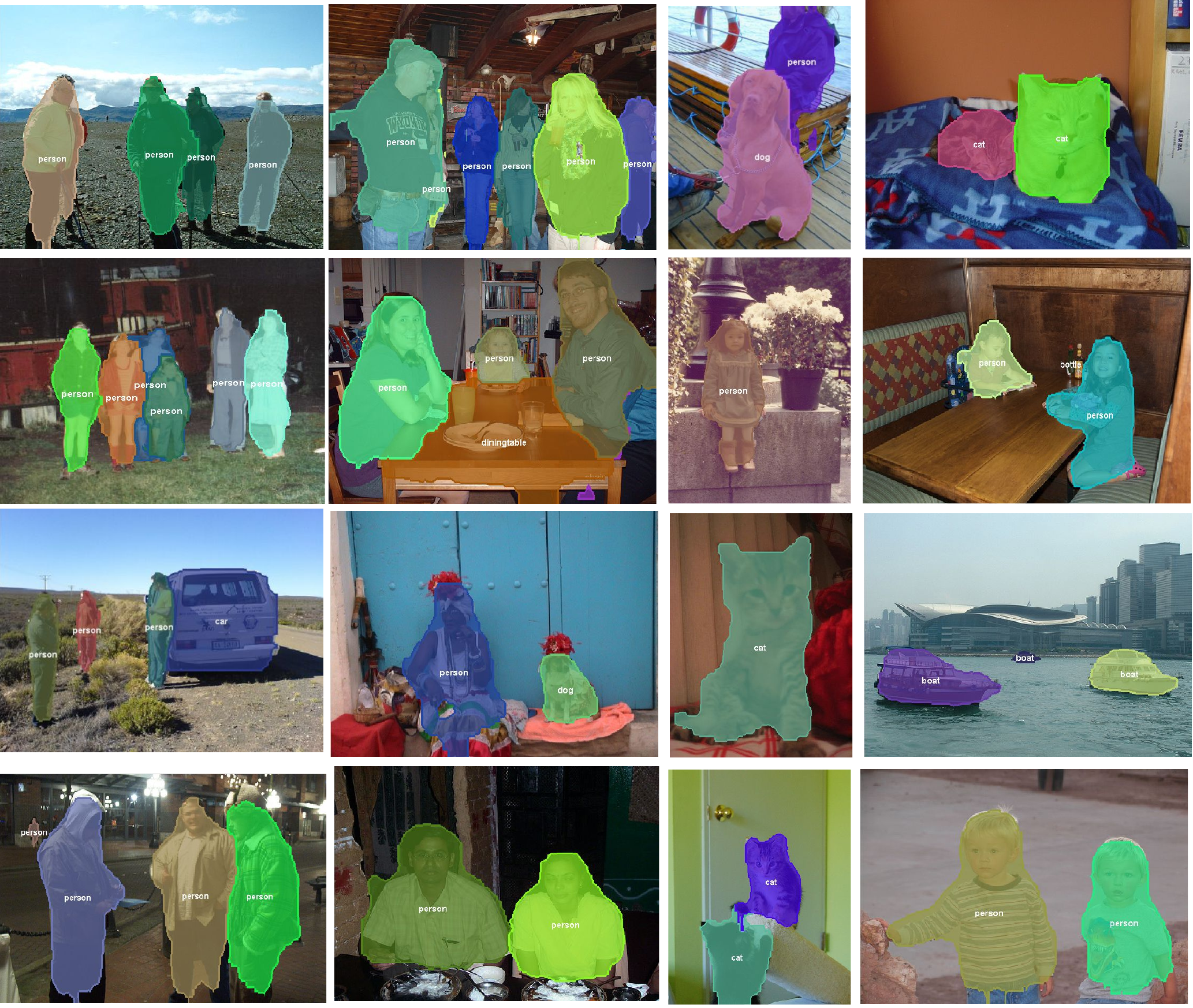}\vspace{-2mm}
\caption{Example object segmentation results of our approach on SDS validation set}\vspace{-4mm}
\label{fig.sds_test}
\end{figure*}

\subsection{Segmentation and Detection on VOC 2012 SDS}

The Simultaneous Detection and Segmentation (SDS) task is widely used to evaluate instance-aware segmentation methods~\cite{hariharan2014simultaneous,dai2015instance}. Following the protocols in~\cite{hariharan2014simultaneous,dai2015instance}, the model training and evaluation are performed on 5623 images from VOC 2012 \emph{train}, and 5732 images from VOC 2012 \emph{validation} sets, respectively. The ground-truth instance-level segmentation masks are provided by the additional annotations from~\cite{hariharan2011semantic}. The mask-level mAP$^r$ scores and the box-level mAP$^b$ scores are employed as the evaluation metrics for instance-level mask estimation and object detection performance measure, respectively.

First, we report the upper-bound result of training the PAD model using ground-truth object masks, i.e., PAD {\scriptsize w/ gt mask} in Table~\ref{tab:voc_sds}). It achieves 68.1\% mAP$^b$  and 64.5\% mAP$^r$. We note that this ``oracle" upper bound is strong. The segmentib
ation accuracy is even higher than the state-of-the-art instance-aware segmentation method, MNC~\cite{dai2015instance}.

Second, we evaluate the superiority of alternative training for pseudo mask estimation and network. As shown in Table~\ref{tab:voc_sds}, PAD obtains mAP$^b$ 67.6\% and mAP$^r$ 58.5\%, which are slightly worse than the upper bound. The instance-level object segmentation results steadily improve with more iterative refinements (iter.1, iter.2). The iterative improvement can be observed in Figure~\ref{fig.graph_cut_example_result}. Finally, PAD {\scriptsize w/ Grabcut mask} shows the instance-level object segmentation results by using the pseudo masks obtained by Grabcut method~\cite{rother2004grabcut}, which is a traditional state-of-the-art object segmentation method. PAD {\scriptsize w/o 1D box loss} corresponds to results without using the 1D box loss as in Eq.~\ref{eq:loss_seg}. Their comparison with PAD in Table~\ref{tab:voc_sds} demonstrate the effectiveness of iterative graph cut refinement and 1D segmentation loss, the key indigents of PAD. Examples for object detection and segmentation are shown in Figure~\ref{fig.test_det} and~\ref{fig.sds_test}.
\subsection{Complexity Analysis}
On average, under ResNet-101, our Faster RCNN baseline
requires 1.5 sec to process each image, during training. The
PAD increases this to 1.9 sec. During testing, the baseline
requires 0.42 sec and the PAD 0.49 sec. Note that
the overall training time is non-trivially larger, due to the
iterative training, but this has no impact in testing. The learned
network is only slightly slower than the Faster RCNN. Regarding
parameters (see Figure~\ref{fig.net1}), the PAD has 3 additional
$1 \times 1$ conv layers (1 in InstanceFCN, 1 for global feedback, 1
for instance feedback). These add about 1M parameters. For
reference, VGG has 134M and ResNet-101 has 42M. The
increased capacity can be considered minor.
\section{Conclusion}
In this work, we present a novel Pseudo-mask Augmented object Detection (PAD) model to facilitate object detection with the instance-level segmentation information that are only supervised by bounding box annotation. Starting from the joint object detection and instance segmentation network, the proposed PAD recursively estimates the pseudo ground-truth object masks from the instance-level object segmentation network training, and then enhance the detection network with a top-down segmentation feedback.
\vspace{-10mm}
\subsubsection*{Acknowledgements}
Thanks Jifeng Dai the help and discussion for this work.
{\small
\bibliographystyle{ieee}
\bibliography{pseudo_mask}

\begin{thebibliography}{10}\itemsep=-1pt

\bibitem{bell2015inside}
S.~Bell, C.~L. Zitnick, K.~Bala, and R.~Girshick.
\newblock Inside-outside net: Detecting objects in context with skip pooling
  and recurrent neural networks.
\newblock In {\em CVPR}, 2016.

\bibitem{boykov2004graphcut}
Y.~Boykov and V.~Kolmogorov.
\newblock An experimental comparison of min-cut/max-flow algorithms for energy
  minimization in vision.
\newblock {\em TPAMI}, 2004.

\bibitem{boykov2001interactive}
Y.~Y. Boykov and M.-P. Jolly.
\newblock Interactive graph cuts for optimal boundary and region segmentation
  of objects in nd images.
\newblock In {\em ICCV}, 2001.

\bibitem{carreira2010constrained}
J.~Carreira and C.~Sminchisescu.
\newblock Constrained parametric min-cuts for automatic object segmentation.
\newblock In {\em Computer Vision and Pattern Recognition (CVPR), 2010 IEEE
  Conference on}, pages 3241--3248. IEEE, 2010.

\bibitem{chen2014enriching}
X.~Chen, A.~Shrivastava, and A.~Gupta.
\newblock Enriching visual knowledge bases via object discovery and
  segmentation.
\newblock In {\em Proceedings of the IEEE conference on computer vision and
  pattern recognition}, pages 2027--2034, 2014.

\bibitem{dai2016instance}
J.~Dai, K.~He, Y.~Li, S.~Ren, and J.~Sun.
\newblock Instance-sensitive fully convolutional networks.
\newblock In {\em ECCV}, 2016.

\bibitem{dai2015boxsup}
J.~Dai, K.~He, and J.~Sun.
\newblock Boxsup: Exploiting bounding boxes to supervise convolutional networks
  for semantic segmentation.
\newblock In {\em ICCV}, 2015.

\bibitem{dai2015instance}
J.~Dai, K.~He, and J.~Sun.
\newblock Instance-aware semantic segmentation via multi-task network cascades.
\newblock In {\em CVPR}, 2015.

\bibitem{dai2016rfcn}
J.~Dai, Y.~Li, K.~He, and J.~Sun.
\newblock R-fcn: Object detection via region-based fully convolutional
  networks.
\newblock In {\em NIPS}, 2016.

\bibitem{deng2009imagenet}
J.~Deng, W.~Dong, R.~Socher, L.-J. Li, K.~Li, and L.~Fei-Fei.
\newblock Imagenet: A large-scale hierarchical image database.
\newblock In {\em CVPR}, 2009.

\bibitem{dong2014towards}
J.~Dong, Q.~Chen, S.~Yan, and A.~Yuille.
\newblock Towards unified object detection and semantic segmentation.
\newblock In {\em ECCV}. 2014.

\bibitem{everingham2010pascal}
M.~Everingham, L.~Van~Gool, C.~K. Williams, J.~Winn, and A.~Zisserman.
\newblock {The PASCAL Visual Object Classes (VOC) Challenge}.
\newblock {\em IJCV}, 2010.

\bibitem{felzenszwalb2004efficient}
P.~F. Felzenszwalb and D.~P. Huttenlocher.
\newblock Efficient graph-based image segmentation.
\newblock {\em IJCV}, 2004.

\bibitem{fidler2013bottom}
S.~Fidler, R.~Mottaghi, A.~Yuille, and R.~Urtasun.
\newblock Bottom-up segmentation for top-down detection.
\newblock In {\em Proceedings of the IEEE Conference on Computer Vision and
  Pattern Recognition}, pages 3294--3301, 2013.

\bibitem{gidaris2015locnet}
S.~Gidaris and N.~Komodakis.
\newblock Locnet: Improving localization accuracy for object detection.
\newblock {\em arXiv preprint arXiv:1511.07763}, 2015.

\bibitem{gidaris2015object}
S.~Gidaris and N.~Komodakis.
\newblock Object detection via a multi-region \& semantic segmentation-aware
  cnn model.
\newblock {\em arXiv preprint arXiv:1505.01749}, 2015.

\bibitem{gidaris2016locnet}
S.~Gidaris and N.~Komodakis.
\newblock Locnet: Improving localization accuracy for object detection.
\newblock In {\em CVPR}, 2016.

\bibitem{girshick2015fast}
R.~Girshick.
\newblock {Fast R-CNN}.
\newblock In {\em ICCV}, 2015.

\bibitem{girshick2014rich}
R.~Girshick, J.~Donahue, T.~Darrell, and J.~Malik.
\newblock Rich feature hierarchies for accurate object detection and semantic
  segmentation.
\newblock In {\em CVPR}, 2014.

\bibitem{gokberk2013segmentation}
R.~Gokberk~Cinbis, J.~Verbeek, and C.~Schmid.
\newblock Segmentation driven object detection with fisher vectors.
\newblock In {\em Proceedings of the IEEE International Conference on Computer
  Vision}, pages 2968--2975, 2013.

\bibitem{hariharan2011semantic}
B.~Hariharan, P.~Arbel{\'a}ez, L.~Bourdev, S.~Maji, and J.~Malik.
\newblock Semantic contours from inverse detectors.
\newblock In {\em ICCV}, 2011.

\bibitem{hariharan2014simultaneous}
B.~Hariharan, P.~Arbel{\'a}ez, R.~Girshick, and J.~Malik.
\newblock Simultaneous detection and segmentation.
\newblock In {\em ECCV}. 2014.

\bibitem{he2017mask}
K.~He, G.~Gkioxari, P.~Doll{\'a}r, and R.~Girshick.
\newblock Mask r-cnn.
\newblock In {\em Computer Vision (ICCV), 2017 IEEE International Conference
  on}, pages 2980--2988. IEEE, 2017.

\bibitem{he2016deep}
K.~He, X.~Zhang, S.~Ren, and J.~Sun.
\newblock Deep residual learning for image recognition.
\newblock In {\em CVPR}, 2016.

\bibitem{jia2014caffe}
Y.~Jia, E.~Shelhamer, J.~Donahue, S.~Karayev, J.~Long, R.~Girshick,
  S.~Guadarrama, and T.~Darrell.
\newblock Caffe: Convolutional architecture for fast feature embedding.
\newblock {\em arXiv preprint arXiv:1408.5093}, 2014.

\bibitem{khoreva2017simple}
A.~Khoreva, R.~Benenson, J.~Hosang, M.~Hein, and B.~Schiele.
\newblock Simple does it: Weakly supervised instance and semantic segmentation.
\newblock In {\em Proc. CVPR}, 2017.

\bibitem{koltun2011efficient}
V.~Koltun.
\newblock Efficient inference in fully connected crfs with gaussian edge
  potentials.
\newblock In {\em NIPS}, 2011.

\bibitem{kong2017ron}
T.~Kong, F.~Sun, A.~Yao, H.~Liu, M.~Lu, and Y.~Chen.
\newblock Ron: Reverse connection with objectness prior networks for object
  detection.
\newblock In {\em IEEE Conference on Computer Vision and Pattern Recognition},
  volume~1, page~2, 2017.

\bibitem{kong2016hypernet}
T.~Kong, A.~Yao, Y.~Chen, and F.~Sun.
\newblock Hypernet: Towards accurate region proposal generation and joint
  object detection.
\newblock {\em arXiv preprint arXiv:1604.00600}, 2016.

\bibitem{lafferty2001conditional}
J.~Lafferty, A.~McCallum, and F.~Pereira.
\newblock Conditional random fields: Probabilistic models for segmenting and
  labeling sequence data.
\newblock In {\em Proceedings of the eighteenth international conference on
  machine learning, ICML}, volume~1, pages 282--289, 2001.

\bibitem{li2017fully}
Y.~Li, H.~Qi, J.~Dai, X.~Ji, and Y.~Wei.
\newblock Fully convolutional instance-aware semantic segmentation.
\newblock In {\em IEEE Conf. on Computer Vision and Pattern Recognition
  (CVPR)}, pages 2359--2367, 2017.

\bibitem{li2004lazy}
Y.~Li, J.~Sun, C.-K. Tang, and H.-Y. Shum.
\newblock Lazy snapping.
\newblock In {\em ACM Transactions on Graphics (ToG)}, volume~23, pages
  303--308. ACM, 2004.

\bibitem{lin2016scribblesup}
D.~Lin, J.~Dai, J.~Jia, K.~He, and J.~Sun.
\newblock Scribblesup: Scribble-supervised convolutional networks for semantic
  segmentation.
\newblock In {\em CVPR}, 2016.

\bibitem{lin2017focal}
T.-Y. Lin, P.~Goyal, R.~Girshick, K.~He, and P.~Doll{\'a}r.
\newblock Focal loss for dense object detection.
\newblock {\em arXiv preprint arXiv:1708.02002}, 2017.

\bibitem{liu2015semantic}
Z.~Liu, X.~Li, P.~Luo, C.-C. Loy, and X.~Tang.
\newblock Semantic image segmentation via deep parsing network.
\newblock In {\em Proceedings of the IEEE International Conference on Computer
  Vision}, pages 1377--1385, 2015.

\bibitem{najibi2015g}
M.~Najibi, M.~Rastegari, and L.~S. Davis.
\newblock G-cnn: an iterative grid based object detector.
\newblock {\em arXiv preprint arXiv:1512.07729}, 2015.

\bibitem{pinheiro2015learning}
P.~O. Pinheiro, R.~Collobert, and P.~Dollar.
\newblock Learning to segment object candidates.
\newblock In {\em NIPS}, 2015.

\bibitem{ren2015faster}
S.~Ren, K.~He, R.~Girshick, and J.~Sun.
\newblock {Faster R-CNN: Towards real-time object detection with region
  proposal networks}.
\newblock In {\em NIPS}, 2015.

\bibitem{rother2004grabcut}
C.~Rother, V.~Kolmogorov, and A.~Blake.
\newblock Grabcut: Interactive foreground extraction using iterated graph cuts.
\newblock {\em ACM Transactions on Graphics}, 2004.

\bibitem{russakovsky2015imagenet}
O.~Russakovsky, J.~Deng, H.~Su, J.~Krause, S.~Satheesh, S.~Ma, Z.~Huang,
  A.~Karpathy, A.~Khosla, M.~Bernstein, et~al.
\newblock Imagenet large scale visual recognition challenge.
\newblock {\em International Journal of Computer Vision}, 115(3):211--252,
  2015.

\bibitem{shotton2006textonboost}
J.~Shotton, J.~Winn, C.~Rother, and A.~Criminisi.
\newblock Textonboost: Joint appearance, shape and context modeling for
  multi-class object recognition and segmentation.
\newblock In {\em European conference on computer vision}, pages 1--15.
  Springer, 2006.

\bibitem{shrivastava2016training}
A.~Shrivastava, A.~Gupta, and R.~Girshick.
\newblock Training region-based object detectors with online hard example
  mining.
\newblock In {\em CVPR}, 2016.

\bibitem{simonyan2014very}
K.~Simonyan and A.~Zisserman.
\newblock Very deep convolutional networks for large-scale image recognition.
\newblock {\em arXiv preprint arXiv:1409.1556}, 2014.

\bibitem{simonyan2015very}
K.~Simonyan and A.~Zisserman.
\newblock Very deep convolutional networks for large-scale image recognition.
\newblock In {\em ICLR}, 2015.

\bibitem{szegedy2015going}
C.~Szegedy, W.~Liu, Y.~Jia, P.~Sermanet, S.~Reed, D.~Anguelov, D.~Erhan,
  V.~Vanhoucke, and A.~Rabinovich.
\newblock Going deeper with convolutions.
\newblock In {\em CVPR}, 2015.

\bibitem{yang2016craft}
B.~Yang, J.~Yan, Z.~Lei, and S.~Z. Li.
\newblock Craft objects from images.
\newblock {\em arXiv preprint arXiv:1604.03239}, 2016.

\bibitem{zheng2015conditional}
S.~Zheng, S.~Jayasumana, B.~Romera-Paredes, V.~Vineet, Z.~Su, D.~Du, C.~Huang,
  and P.~Torr.
\newblock Conditional random fields as recurrent neural networks.
\newblock In {\em ICCV}, 2015.

\end{thebibliography}
}

\end{document}